# 格数致知：走向对世界的深度认知

胡包钢[1, 2]　董未名[1, 2]

**摘　要**　"格数致知"是我们集成东西方哲学家思想而发展出的新提法。它可以具体解释为："推究事物之内在本源，获取知识之数学表达"。本文是基于人工智能发展背景下对这个学术思想展开探讨的观点文章。从机器学习研究视角讨论了智能机器最终发展目标，以及两种知识表达及其与"格数致知"的关联。在给出客观评价元准则方法论之后，针对人工智能在国家发展或治理中一个客观综合评价应用实例说明"格数致知"思想的广泛应用前景及其可能存在的问题。

**关键词**　格数致知, 人工智能, 知识, 表达，社会计算，客观评价

## "Ge Shu Zhi Zhi": Towards Deep Understanding about Worlds

HU Bao-Gang[1, 2]　DONG Wei-Ming[1, 2]

**Abstract**　"Ge Shu Zhi Zhi" is a novel saying in Chinese, stated as "To investigate things from the underlying principle(s) and to acquire knowledge in the form of mathematical representations". The saying is adopted and modified based on the ideas from the Eastern and Western philosophers. This position paper discusses the saying in the background of artificial intelligence (AI). Some related subjects, such as the ultimate goals of AI and two levels of knowledge representations, are discussed from the perspective of machine learning. A case study on objective evaluations over multi attributes, a typical problem in the filed of social computing, is given to support the saying for wide applications. A methodology of meta rules is proposed for examining the objectiveness of the evaluations. The possible problems of the saying are also presented.

**Key words**　Ge Shu Zhi Zhi, Artificial Intelligence, Knowledge, Representation, Computation Social Science, Objective Evaluation

## 1 引言

人类文明发展的进化史是根据应用工具中的材料来划分的。人类创造工具，工具改变人类。人工智能作为一种新的工具已经远远超出传统工具的内涵，智能机器会被赋予更多人类功能属性(比如对话机器人中包含价值观或文化背景)，它对人类未来生活方式将会带来更为深刻的改变。因此人工智能学科大大超越其它学科拥有的基础原理范围，涉及到哲学、人文学、社会学、认知科学、心理学、语言学、知识科学、信息科学、计算科学、工程科学等诸多学科的内容[1-3]。

本文是针对人工智能研究发展目标中学术思想观点讨论文章。面对这样的主题无法避免要有哲学思想视角讨论。Duda等人在他们人工智能经典教材《模式分类》中指出，"那是什么？（What is that？）"同样"是哲学知识论所面临的中心问题：即探寻知识的本质。(It has been a central theme in the discipline of philosophical epistemology, the study of the nature of knowledge)"[4]。考虑人工智能宏大的学科内涵，本文限定从"哲学思想与计算建模"的关联视角予以探讨。这是由于人工智能必然要落实到计算模型为实体[5]，这种计算层面视角可以直接导向许多问题的核心与本质。在借鉴东西方哲学思想之后，我们提出了"格数致知"的学术思想。第2节对"格数致知"提法予以具体说明。第3节从机器学习研究视角讨论该提法。第4节给出了一个具体应用实例。第5节总结了"格数致知"提法的意义及可能的问题。

需要指出本文涉及的一些术语（如数学、知识、表达、计算、人工智能等）至今学界没有发展出统一定义。因此我们采取直接应用术语方式，并给予广义方式理解。比如将"数学化"术语包容到"公理化"和"形式化"的内涵[6]。本文重点是强调学术思想层面新见解提出，期望激发不同领域研究者质疑与创新。

## 2 关于"格数致知"提法的说明

本文借鉴东西方哲学家学术思想提出"格数致知"说法，目标是继承与发展前人学术思想。为此我们采取吸取先贤思想中积极内涵与现代理解的立场来解释有关学术观点，而不是原有思想本意的考证。第一位是中国先哲老子(约公元前571年-约前471年)提出的"道生万物"。该思想的原文是："道生一，一生二，二生三，三生万物"。老子的观点包含诸多智慧要点。其中有宇宙万物演变而来、由简至繁、自有规律、"道"乃唯一源泉。可以理解"道"或"源"也

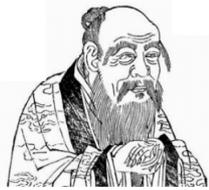 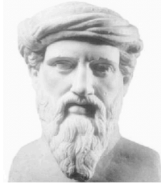 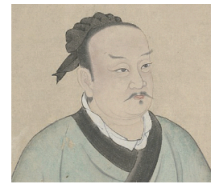

老子
(約公元前 571-471)
道生萬物
All originates from the Tao.

畢達哥拉斯
（約公元前 570-495）
萬物皆數
All is number.

曾子
（公元前 505-435）
格物致知
Investigating things and acquiring knowledge.

格數致知
簡要解釋：推究事物之內在本源
獲取知識之數學表達
To investigate things from the underlying principle(s)
To acquire knowledge in the form of mathematical representations

图 1: 关于"格数致知"新提法的起源（图片来自网站）。
Fig 1 The sources of the new saying about "Ge Shu Zhi Zhi"（Figures are taken from web sources）.

成为科学家追求科学研究的目标之一。爱因斯坦（Albert Einstein）曾终生致力于发展物理世界中的统一理论[7]。他特别强调"自然乃为最简洁且可理解的数学思想之实现（Nature is the realization of the simplest conceivable mathematical ideas）"[8]。

第二位是古希腊哲学家毕达哥拉斯(约公元前570-约前495)提出的"万物皆数"的学术思想。从现代意义上可以理解是将"数"归结为万物本源（尽管毕达哥拉斯原有认知是限于有理数范畴内）。对于客观世界中实体万物的划分及其时空变化是离不开"数"的表达来体现的。这种推论同样适用于主观世界中的虚拟万物。只有应用"数"方能辨识和解释所有万物，这也成为人能够"感知"或"认知"万物的基础。"数"为万物本源的另一个理解是"数"乃独立于并先于万事万物。由此产生了"数与道"之间的本源问题。

第三是出自中国《礼记·大学》篇中提出的"格物致知"思想，相传为曾参(公元前505-公元前435)所作。它可以解释为：探察事物，获取知识。中国历史上曾应用"格致"一词表示研究事物的学问。早期也曾用于英文"Science"的翻译，之后被"科学"术语所取代[9]。从人文或科学背景理解的重要内涵有"致知"应为人生或科学目标，"格物"方是必然路径。

综合上述东西方哲学家学术思想，"格数致知"提法可以简要解释为"推究事物之内在本源, 获取知识之数学表达"（图1）。"之"字可以按"的"意理解。解释语中反映任何事物发生与发展都有其规律与原理，它们揭示了事物的内在本源。另一方面，知识获取中强调知识深层理解主要取决于事物规律与原理的数学语言表达。其中我们将知识大体分为两个层面：浅层知识与深层知识。由此也对应了两种语言方式来表达，分别是自然语言与数学语言。牛顿定律就是一个"格数致知"的最好实例。表1示意了该实例中关于知识层次与语言表达的基本关联。在下节中我们还会讨论自然语言与数学语言在表达知识方面中的具体差异。

表 1 关于知识层次与语言表达的基本关联及其实例
Table 1 An example of the relations between knowledge levels and representation types.

| 知识层次 | 语言方式 | 实例 |
| --- | --- | --- |
| 浅层知识 | 主要为自然语言表达 | 外力越大物体的加速度越大 |
| 深层知识 | 主要为数学语言表达 | $\boldsymbol{F} = m\boldsymbol{a}$ |

## 3 基于机器学习研究视角的理解

目前的机器学习主要是基于数据驱动、以归纳为推断原理而实现的一种人工智能方法。本文是从机器学习研究视角来讨论人工智能以及"格数致知"的说法，期待能够给出独到见解。文[10]中将机器学习按照四个基本问题展开讨论（图2）。它们分别是：

基本问题一："学什么？"，也可称为"学习目标选择"。"学习目标"是对机器模型预期学习内容的"表达（Representation）"。该表达（又可称为表示）可以分解为两个层次：即"语言（Linguistic）"表达与"计算（Computational）"表达。它们分别对应为应用自然语言的定性叙述与应用计算语言的定量描述。在计算表达层面通常要涵盖应用数学语言描述的目标函数，约束条件，以及优化形式

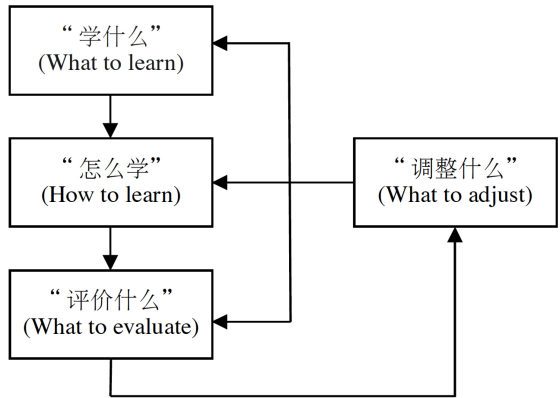

图 2: 机器学习研究中基本问题(或层次)流程示意图 [10]。

Fig 2 Four basic problems (or levels) in machine learning [10].

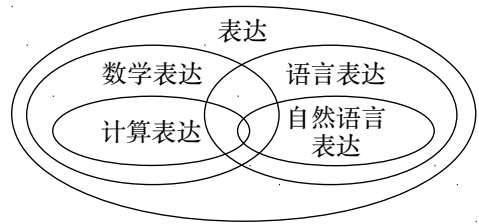

图 3: 关于表达与其它方式表达关联的文氏图。

Fig 3 A Venn diagram of representations in relation to the other forms.

三个方面的内容。该专题研究直接涉及了语言学、心理学、认知科学、信息科学、以至人文、社会学的领域知识，强调机器学习对学习机制的深刻认知，包括语言表达与计算表达两个层面中显式知识的获取。

基本问题二："怎么学？"。该专题主要应用统计学、优化、计算理论、以及量子计算为数学理论基础。广义理解应该包括工程实现，大量工程基础知识为其应用基础。

基本问题三："评价什么？"又可称"评价指标（或度量函数）选择"。当设定目标函数与评价指标应用完全相同的数学函数表达时，该问题成为基本问题一中的子问题（无需选择优化形式）。目前实际应用中常会采用单一学习目标，而以多项指标对其进行综合评价的方式。多指标与学习目标通常情况下并非具有一致性，因此"评价什么"将成为基本问题。

基本问题四："调整什么？"是关于前三个模块动态调整的专题研究。专题目标是赋予机器具有"智能进化"功能。该进化包含了知识层面中显式更新方式。

图2中基本问题关联图示意了对复杂系统实施一种"兼容并包"哲学思想的研究方式：即要有"分而治之"的"还原论(Reductionism)"，又要兼容"合而为一"的"整体论(Holism)"哲学思想。它大体符合人的智能行为与决策习惯。从机器设计流程角度讲，"学什么？"与"评价什么？"一起构成了首要问题，"怎么学？"是次要问题，"调整什么？"是第三级问题[10]。然而次级问题也会对上一级问题产生影响，比如"怎么学？"中高效计算设计会要求"学什么？"中学习目标函数可能更改为凸函数来近似。图2中基本问题划分强调了有关"知识"在"计算"层面中的转换与关联，它有助于我们梳理机器学习研究中的科学问题与工程问题。

我们应用文氏图（Venn diagram）方式示意出"表达"中多种不同表达方式及其之间的关联(图3)。要理解图中省略了其它表达方式，如图表达与音乐表达等。人工智能中"语义鸿沟(Sematic Gap)"的挑战性问题正是发生在自然语言表达与计算表达（或语言表达与数学表达）之间的知识转换。当自然语言表达转换为计算表达时，我们通常遇到的数学问题是"病态定义（Ill-defined）"形式，典型实例有个人情感（如高兴）状态描述。而从计算表达转换为自然语言表达的过程中，除了原有"病态定义"问题仍然会出现外，另一个数学问题"不适定（Ill-posed）"形式也能更多遇到。典型实例有应用二维数据表达来近似真实三维物体表达。为此，"语义鸿沟"中可以有以下两个推论：

推论一. 自然语言表达与计算表达通常没有"一对一"关系。

推论二. 在惟一语义理解下，计算表达到自然语言表达有"一对一"关系。

推论二的实例有牛顿第二定律中应用自然语言表达计算公式。推论的重要意义是：知识理解中"数学表达"方具惟一性，"语言表达"存有歧义性。这也就是为什么"格数致知"中我们强调"数学表达"而非"语言表达"的缘故。由于"语言表达"的知识通常会有"见仁见智"或"浅尝辄止"的解释，因此相对而言又可将"数学表达"的知识称为深度知识。应用"数学表达"是为了保证人们能够对知识被完整以及惟一方式的理解。而人工智能或机器学习的最终发展目标可以是[11]：

目标一. 追求最大效率地利用智能机器为人类服务。

目标二. 追求人类对各种世界的深度认知。

由此又可以分别称为"工具论"目标与"认知论"目标。在"工具论"目标中实际上隐含了"认知论"目标。因为对工具开发者而言，工具的改进与完

善需求对工具自身的深度理解。而对用户而言，傻瓜式或黑箱式操作是工具的需求特征。"认知论"目标中也隐含了对工具的需求：利用数学工具来获得深度认知。"认知论"目标中我们将面临的最大挑战有：

挑战一. 是否有哪类知识是不可认知的，或不可计算的？

如果采纳波普三种世界（物理世界、精神世界和客观知识世界）说法[12]来讨论，我们知道精神世界中存在"只可意会，不可言传"的事物，比如某种意识活动。如果确实无法用语言表达这种活动，可以说我们将无法获得相关的知识。这也意味"格数致知"并非肯定人类能够认知全部世界，不可认知的边界本身就是"格数致知"中的重要主题之一，其意义犹如揭示认知活动或机器智能中的永动机。"格数致知"的提法将引导我们从数学层面去探讨机器智能能否全面超越人类智能，比如以计算表达或可计算性的角度来解释。

## 4 一个具体应用实例

本节期待通过世界各国生活质量评价的具体实例说明"格数致知"的意义及其实施办法。这个实例属于人工智能在社会科学中的具体应用。它也可以理解为是社会计算或社会智能的研究内容[13]。人工智能为全球、国家、以及地区发展或治理提供了全新的解决方法。

我们仍然借鉴图2中机器学习中基本问题划分方式讨论社会发展为主题的应用背景。因此四个基本问题分别是："发展什么？"，"怎样发展？"，"评价什么？"，"调整什么？"。本实例中设定生活质量为发展目标。可以理解这是多目标优化问题。按照"格数致知"的思想，需要对四个基本问题分别予以语言表达与计算表达的具体定义。本文仅对生活质量中"评价什么？"给予示例讨论，并认为这是社会发展或治理中的首要问题。因为"评价什么？"可能会成为实际社会发展中的指挥棒，而非"发展什么？"中的发展目标设定。"评价什么？"同样可以用于"怎样发展？"与"调整什么？"中的实施效果考察。从机器学习角度看，由于无法给出综合评价（如各国生活质量）标准答案，多指标综合评价是无监督学习问题。为跨越该类问题中的难关，我们提出了"元准则"评判方法。下面首先给出元准则定义：

元准则是关于具体评价准则之上的一组高层知识准则（为一个集合），主要用于评判具体评价准则是否符合规定的高层知识。

元准则是人工赋予的准则，它并非代表正确或一致。根据问题背景具体元准则会发生变化。元准则评判方法就是应用元准则实现评价的一种方法。下面通过世界各国生活质量客观综合评价实例给予具体讨论。可以理解评价或推理本身属于主观世界产物。但是区分客观还是主观评价是人们生活中的基本需求，也是评价应用中需要首先明确的问题。Berger从哲学视角列出了需求客观性的四点原因[14]。"让数据说话"的客观评价可以成为主观评价的参照基础但是更具难度。然而，学界缺失关于"客观性"在技术层面上予以检查的方法论。为此我们提出元准则方法论来实现或考察多指标客观综合评价（表2）。可以看到，该表试图从技术层面具体规范或界定评价属性。当评价满足表中元准则后，我们可以归为客观评价，否则归为主观评价。表中元准则并非正确、惟一、完整。它是为避免鸡与鸭讲式对话提供了一种解决方案：建立共识基础（即共同语言）之后实施讨论。因此元准则方法论同样适用于主观评价。综合评价中主观评价（如问卷调查或专家打分）同样是重要的。但是应用中应该明确告知用户有关评价工具的元准则，以及其中主客观成分的各自说明。

表 2 关于实现多指标客观综合评价中的元准则

Table 2 Meta measures in realizing an objective evaluation from multi-attribute objects.

| 考察内容 | 具体元准则 | 说明 |
| --- | --- | --- |
| 指标 | 给定"显式"价值目标，指标选取的依据（如相关性、单调性、可度量性、综合性等） | 指标也可以称为属性、变量、特征、指数、因子等 |
| 数据 | 公开的数据、采样方法、采样规模、无偏性质、归一化方式等内容 | 无偏性质是指不能包容个人打分数据类型 |
| 评价函数 | 给定的输入-输出之间数学关系，评价曲线合理性（如尺度与平移不变性、严格单调性、线性/非线性的兼容性、光滑性、参数规模确定性），无自由参数或自由参数由数据惟一确定 | 评价函数（或曲线）的数学表达：$y = f(\boldsymbol{x}, \boldsymbol{\theta})$ $y$：评价结果输出 $\boldsymbol{x}$：多指标输入向量 $\boldsymbol{\theta}$：参数向量 $f$：多输入单输出函数 |
| 计算机代码 | 公开的算法与代码，计算结果具有可重复性 | 目标于他人可以检查并重复实验 |
| 其它 | 罗列可能的人为因素或适用范围（如非线性类型） | 明确可能的主观性来源或模型的局限性 |

元准则方法论来源于元分析或元评价思想，但是强调应用元准则来降低评价设计中的模糊空间。比如"显式"价值观设定后方可选取指标。元准则选择乃是评价分析中的核心问题。

当以"生活质量"为世界各国评价中的"显式"价值目标时，我们采纳文[15]中应用4项指标来实

现客观综合评价。它们分别是：人均GDP、人均预期寿命（LEB）、婴儿死亡率（IMR）、肺结核率（Tub）。他们研究中首次选择主曲线方法的思路十分具有启迪性。只是在应用"弹性图（Elmap）"方法中未对评价函数施加必要的元准则约束。为完善该工具，我们总结出综合评价中进一步的数学问题是：

数学问题一. 在无标准答案下如何考察评价结果具备合理性？

数学问题二. 引入何种非线性评价函数？

文[16]中给出上述问题解决方案：排序主曲线（RPC）方法（又可称RPC评价准则）。针对数学问题一提出评价曲线合理性的五条元准则（参见表2内容）。比如其中尺度不变性意指排序结果与指标量纲（人民币或美元）无关。读者可以阅读原文中设定元准则的具体原因讨论。可以理解我们是将合理性转化为数学层面上的约束。只有满足五条约束的评价工具方为合理（符合特定的定义），然后可以相互比较（如平均误差）。解答数学问题二中我们选用了三次Bezier曲线为评价函数，因为它在满足合理性的五条元准则情况下能够提供最大非线性变化范围指数（Nonlinearity Variation Index）以及多种简洁性，如归一化、递归性、并行性、参数扩展性与几何直观理解性等特征[17]。新方法中应用可视化图形表达则是强调大数据分析中"让数据说话"与"一目了然"(快速获取知识)的本意。

应用文[15]中公开数据，图4给出了2005年171个国家4项指标数据的直方图、散点图，其中每个绿点为一个国家的数据点（经过归一化处理），红色曲线是应用RPC方法生成的一条排序主曲线在相关二维坐标中的投影。排序主曲线犹如全部数据点的中央骨架，是主成分分析法中"第一主成分的非线性推广[18]"。又可以称它为评价主轴，一端代表生活质量高，另一端代表生活质量低。171个国家数据点正交投影在该主轴后，就能够反映先后排序关系。RPC方法输出排序数据为实数类型，包含更多信息。可以看到171个国家数据点呈现了明显的结构。所谓数据挖掘就是从数据中找出结构性的规律给予具体应用中的知识解释。图4中反映的规律是有些维度之间是单调递增，有些是单调递减。排序主曲线在二维关系中总体呈现简单非线性结构，但是也有近似线性关联的二维关系。因此线性工具如主成分分析法或熵权法无法刻画如此非线性结构。文[16]中考虑了C型、反C型、S型、反S型共四种简单非线性类型，同时RPC方法可以实现全维度或部分维度线性关系的刻画。评价函数自由参数总数为$4 \times d$，$d$是指标维度。RPC方法类似于回归方法通过数据来惟一确定全部自由参数（然而应用正交投

影距离计算误差）。由于没有人为调整参数（如权重），因此新方法更为合理地实现了客观综合评价。此外，自由参数具有归一化与几何直观解释与理解性质。理解性质是指给定自由参数值后，用户可以直接判断非线性类型。图5罗列出2005年171个国家4项指标数据以及应用RPC方法计算的部分国家排序结果与自由参数$\theta$。全部数据与计算代码见[19]，开发者李纯果的研究还包括世界大学排名与计算机科学刊物排名的实例评价内容。

我们在文[16]中明确给出了应用RPC评价准则中非线性类型以及指标内容反映结论的客观性范围（或该研究中的局限性）。从大数据角度看，添加更多指标可以增加综合性与客观性程度，比如正面指标入学率、就业率，负面指标谋杀犯罪率、基尼系数等。由此可见客观（或主观）评价具有相对属性，包容不同程度定义。目前综合评价分析中国际常规是采用加权平均法，如世界大学学术排名(上海交通大学发布的ARWU)、全球竞争力指数排名(世界经济论坛发布的GCI)、全球民主指数排名(经济学人智库发布的DI)等。然而应用个人打分数据(DI)还是非个人打分统计数据(ARWU)在客观性方面呈本质差异。后者更符合客观性原理，而前者反映了打分人的主观偏见。其中公开打分人人数统计数据也是任何综合评价中应该遵循的元准则（DI排名没有公开这方面数据）。

提出元准则方法论也是目标于避免评价工具的误导或滥用。防止人类成为工具下的奴隶将是人工智能发展中的重要研究内容。不管人们喜欢还是不喜欢这类排序工具，日常生活中已是无法离开它们。如人们每天上网查询信息的活动正是应用了排序工具。考察身体状况是健康还是亚健康也是如此。个人或企业都是在选择（包含自我评估）中前行。评价分析是人们认知现状、发现问题、正确决策的一般性前提。本应用实例说明大数据分析或人工智能能够为我们带来知识发现。客观性分析或评价是走向认知真理的必要条件。"格数致知"中我们必须借助数学工具并不断完善它们。

## 5 结论

在科学发展与人工智能兴起的背景下，人类需要不断地继承先贤人文、科学与哲学思想并赋予新的学术内涵。本文正是这样一种初步尝试：即提出"格数致知"的说法。面对纷繁复杂真实世界的演化、神奇奥妙生物大脑的智能机理，人类可谓知之不多。"格数致知"说法将强化我们对知识发现的追求，强调利用数学工具扩展对各种世界的深度认知，由此来有效地应对人类面临的许多共同挑战。

人工智能为我们带来新的发展契机，中国能够

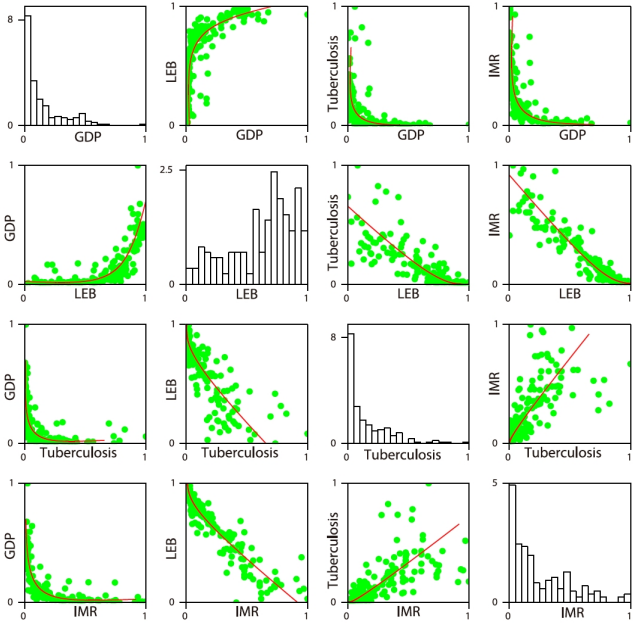

图 4: 2005年171个国家4项指标数据的直方图、散点图及排序主曲线(红色)在二维中的投影[16]。

Fig 4 Histograms, scatter plots, and the projections of the ranking principal curve from RPC for the data in Year 2005, with 171 countries on 4 indexes [16].

而且应该在人类社会进步中走出一条新路：集成东西方智慧、兼容人文价值与科学知识、造福人类命运共同体。马世骏、钱学森等[20,21]分别提出的"社会－经济－自然复合生态系统"、"开放的复杂巨系统"正是体现了中国科学家综合东西方思想发展的重大学术创新。这种基于数学为工具的系统论思想预示我们应该如何认知和治理复杂世界。采用机器学习四个基本问题划分研究框架也可为探索各种世界带来数学层面见解。比如社会发展本应是带约束多目标优化问题（实际约束可能包含未知函数表达而需求机器学习后获得认知），社会博弈中纳什均衡点可能并非是最优解（转化为最优解或次优解的条件会是什么），客观综合评价应视为社会治理中的首要内容之一（如何发展具有共识元准则下的客观评价体系），社会自我修复的反馈机制是否存在或有效（不同国家或地区的反馈机制有哪些并进行比较研究）。多元化学术思想争锋应该提倡，但是应用数学工具方能避免似是而非论断并可跨越见仁见智无效争议陷阱。

此外，本文给出的具体应用实例说明认知世界需要不断发展新的数学工具。比如在多指标综合评价中，元准则方法提供了对客观性程度进行界定或比较的工具，排序主曲线方法则成为考察多维大数据中简单（单调）非线性结构的工具。我们提出了这种新的非线性工具，可以更为有效地表达相关数据的非线性结构，这是传统线性工具无法实现的。应用中我们还应该尽快采纳更为合理的工具，如文[22]在1986年就指出算数平均法在实施单指标排序中的严重缺陷（不满足尺度不变性数学性质），并建议应用几何平均法。然而直到2010年方被国际组织更正为应用几何平均法来计算其人类发展指数（HDI）[23]。"工具改变人类(Tools change humans)"的说法意味着人们行为与社会发展可能与其应用的工具息息相关。相关中正负方面的影响或因果关联同样需求"格数致知"方式的解释。我们需要不断完善工具（如机制设计）来实现人类发展目标。

本文将"格数致知"设为人工智能研究或科学研究以至人类发展目标之一。这一提法及其解释仍需质疑、推敲、以及完善。特别是考虑在应用中可能带来的问题。以下给出三方面内容，目的是尽量减少偏差理解并能够认知相关问题。

相关问题一. "格数致知"是以"格物"而不是"格数"为出发点（纯数学研究也应按事物来理解）。"格数"在"格物"中可以理解为是一种方法论。研究中要同时避免"格物"与"格数"的缺失或脱节。

相关问题二. "格数致知"在人工智能研究中多数是以数学模型来体现。通常情况下，数学模型是趋向获取正确知识的必要条件而非充分条件。然而要理解模型假设与局限性，谨防过度数理化与迷信模型工具以及结论的负面发展趋向。

相关问题三. "格数致知"提法将数学表达设为获得深度知识的基础或必要条件。然而，哥德尔不完备定理表明这个基础本身就有根本性漏洞：完备而自洽的数学体系就不存在。"阿罗不可能定理（Arrow's Impossibility Theorem）"[24]或"辛普森悖论（Simpson's Paradox）"[25]等说明具体应用（如投票选举或医疗方案选择）中的数学困境。

最后一个问题更本质上是"格数致知"提法中基础性问题。上述问题说明应用中要有清醒认识并能够正确实践。"格数致知"并非意指对所有事物都是可以实现"致知"。"知"与"未知"的边界可能是模糊且为动态的，因此该提法更是一种追求目标。如果我们将中国历史百家争鸣中简单分为两大家，则可以有以孔子为代表的"教家(Education School)"（代表作《论语》）和以屈原为代表的"问家(Inquiry School)"（代表作《天问》）。他们分别对应了中国先贤人文教育与科学质疑的两种文化基因。在"格数致知"的道路中，屈原的理念也反映了我们的共同追求：

"路漫漫其修远兮，吾将上下而求索"。

| Country | GDP | LEB | IMR | Tub | Elmap [12] | | RPC | |
|---|---|---|---|---|---|---|---|---|
| | | | | | Score | Order | Score | Order |
| Luxembourg | 70014 | 79.56 | 6 | 4 | 0.892 | 1 | 1.0000 | 1 |
| Norway | 47551 | 80.29 | 3 | 3 | 0.647 | 2 | 0.8720 | 2 |
| Kuwait | 44947 | 77.258 | 11 | 10 | 0.608 | 3 | 0.8483 | 3 |
| Singapore | 41479 | 79.627 | 12 | 2 | 0.578 | 4 | 0.8305 | 4 |
| United States | 41674 | 77.93 | 2 | 7 | 0.575 | 5 | 0.8275 | 5 |
| ⋮ | ⋮ | ⋮ | ⋮ | ⋮ | ⋮ | ⋮ | ⋮ | ⋮ |
| Turkey | 7786 | 71.396 | 13 | 26 | 0.090 | 76 | 0.6298 | 75 |
| Iran | 10692 | 70.618 | 11 | 31 | 0.105 | 69 | 0.6264 | 76 |
| Armenia | 3903 | 73.129 | 32 | 23 | 0.074 | 78 | 0.6248 | 77 |
| China | 4909 | 72.555 | 45 | 21 | 0.079 | 77 | 0.6222 | 78 |
| Samoa | 4872 | 70.807 | 9 | 24 | 0.070 | 81 | 0.6180 | 79 |
| ⋮ | ⋮ | ⋮ | ⋮ | ⋮ | ⋮ | ⋮ | ⋮ | ⋮ |
| $p_0$ | 44713 | 81.218 | 2 | 0 | - | - | - | - |
| $p_1$ | 330 | 80.4 | 2 | 0 | - | - | - | - |
| $p_2$ | 330 | 59.7 | 33 | 43 | - | - | - | - |
| $p_3$ | 1581.824 | 41.68 | 290 | 151 | - | - | - | - |

图 5: 2005年171个国家生活质量评价中部分排序结果[16]。

Fig 5 Part of the ranking list for the life qualities of 171 countries from the data in Year 2005 [16].

说明

这是本文第三版，为最终完整版本。第一版曾于2018年12月19日发布在（https://arxiv.org/abs/1901.01834）。